\documentclass[10pt,twocolumn,letterpaper]{article}

\usepackage{cvpr}
\usepackage{times}
\usepackage{epsfig}
\usepackage{graphicx}
\usepackage{amsmath}
\usepackage{amssymb}
\usepackage{breqn}
\usepackage{mathtools}
\usepackage{booktabs}
\usepackage{verbatim}
\usepackage{tikz}
\usepackage{subcaption}
\usepackage{setspace}
\usepackage{array}

\usepackage[pagebackref=true,breaklinks=true,letterpaper=true,colorlinks,bookmarks=false]{hyperref}
\graphicspath{{images/}}
\cvprfinalcopy 

\def\cvprPaperID{2678} 

\ifcvprfinal\fi
\begin{document}

\title{Deep Stereo Matching with Dense CRF Priors}

\author{
 Ron Slossberg\\ 
 {\tt\small ronslos@tx.technion.ac.il}
 \and
 Aaron Wetzler\\
 {\tt\small twerd@cs.technion.ac.il}
 \and
 Ron Kimmel\\
 {\tt\small ron@cs.technion.ac.il}
 \and
 Technion - Israel Institute of Technology\\
 Haifa, Israel
 }

\maketitle

\begin{abstract}
Stereo reconstruction from rectified images has recently been revisited within the context of deep learning. Using a deep Convolutional Neural Network to obtain patch-wise matching cost volumes has resulted in state of the art stereo reconstruction on classic datasets like Middlebury and Kitti. By introducing this cost into a classical stereo pipeline, the final results are improved dramatically over non-learning based cost models. However these pipelines typically include hand engineered post processing steps to effectively regularize and clean the result. Here, we show that it is possible to take a more holistic approach by training a fully end-to-end network which directly includes regularization in the form of a densely connected Conditional Random Field (CRF) that acts as a prior on inter-pixel interactions. We demonstrate that our approach applied to both synthetic and real world datasets outperforms an alternative end-to-end network and compares favorably to less holistic approaches. 
\end{abstract}

\section{Introduction}
Stereo reconstruction is a well studied problem in the field of computer vision. It is based on the notion that by taking two images of a scene from slightly different angles, one can reconstruct the depth of the scene by triangulation. This is in fact how humans observe the world. However, despite the apparent ease with which we perceive depth, the computational version of the problem is far from trivial. Many approaches have been developed to solve the problem of stereo depth perception and in general they all consist of two main stages: correspondence and regularization. The correspondence stage attempts to find a match for each of the pixels between the stereo images, while the regularization typically applies global constraints such as smoothness priors to the matching process to improve the fidelity and accuracy of the matched locations. 

Convolutional Neural Networks (CNN) are an efficient and successful computer learning tool which have been demonstrated to solve many difficult problems in computer vision. The paper by Zbontar \etal \cite{zbontar2015computing} has recently shown that CNNs can provide state of the art results in stereo reconstruction, providing both accuracy and computational efficiency. That work however applied a neural network to the image correspondence part of the algorithm successfully but did not deal with new regularization techniques, instead relying on existing algorithms. In a similar context Mayer \etal \cite{MIFDB16} also applied a CNN methodology to the stereo matching problem, but performed the disparity prediction in an end-to-end manner without including a specific regularization stage. Their score on the KITTI \cite{Menze2015CVPR} benchmark is somewhat lower than that of \cite{zbontar2015computing} which may be attributed to training with an $L_2$ norm regularizer which is prone to smoothing around object boundaries. We instead propose an end-to-end network which aims to preserve sharper object boundaries while explicitly incorporating a powerful regularization stage in the form of a Conditional Random Field (CRF). Our method uses a mean-field network layer as first shown in \cite{CRFRNN}. Doing so imposes a more natural piecewise smoothness prior, which assumes local smoothness with sparse discontinuities around object borders. The parameters controlling these priors are tunable and are learned directly from examples within the CNN framework. Furthermore we introduce a novel compatibility function for our CRF prior to better represent disparity neighborhood interactions. In this paper we show how this network can be constructed and trained using both real world and synthetic data. We discuss the benefits obtained by including the CRF stage as part of the network and compare our results to those achieved by \cite{MIFDB16,zbontar2015computing}.

\section{Related work}
In recent years large stereo datasets with ground truth disparity such as KITTI \cite{Geiger2012CVPR,Menze2015CVPR} and Middlbury \cite{scharstein2002taxonomy} have been made available to the research community. These real life datasets have been supplemented by much larger synthetic datasets \cite{MIFDB16} which were specifically created to enable deep learning for stereo matching. This has led to a new generation of stereo reconstruction algorithms with model parameters that are tuned from real-world or synthetic data. Stereo reconstruction which is based on supervised learning can be roughly characterized into three main groups which we briefly review here. For a more general treatment of stereo algorithms we refer the reader to \cite{scharstein2002taxonomy}.

{\bf Non-CNN learning based methods.} 
The multiple-experts method of \cite{kong2004method,kong2006stereo} calculates an initial matching cost and then refines it using a classifier which is learned from data. Similarly \cite{peris2012towards} uses AD-Census \cite{mei2011building} to calculate an initial matching cost and then learns the relation between the matching cost and final disparity using LDA. A more recent approach taken by \cite{haeusler2013ensemble} uses a random forest classifier to combine several confidence measures to improve the matching cost. In \cite{spyropoulos2014learning} the authors also use a random forest classifier for matching patches and then refine the result using an MRF model.

{\bf CNN based methods}
CNN based methods have been shown to be effective at computing the matching cost. A deep network was demonstrated in \cite{zagoruyko2015learning} which was trained to evaluate patch similarity. This idea was harnessed by \cite{zbontar2015computing} and \cite{Chen_2015_ICCV} who applied a CNN to calculate the cost for matching two patches taken from stereo pairs. The CNNs are trained on patches taken from the labeled dataset and the locally calculated matching cost is directly used to estimate the final disparity without taking into account global priors during the training process.  In contrast to \cite{zbontar2015computing}, \cite{MIFDB16} has taken an end-to-end deep learning approach to the stereo matching problem. In this work, a deep network is used and trained in a multi-scale end-to-end manner. Additionally, \cite{MIFDB16} created large synthetic datasets for the purpose of training their network. 

{\bf CRF/MRF based methods}
Markov and conditional random fields have long been used to impose global priors onto noisy disparity estimations. An MRF model demonstrated in \cite{zhang2007estimating} learns hyper-parameters from a single stereo pair. Similarly, the work of \cite{scharstein2007learning} shows how constructing a labeled dataset consisting of $30$ stereo pairs can be used to train the parameters of a CRF. An alternative parameter training approach for CRF based denoising is shown in \cite{li2008learning} where the model parameters are determined using a structured support vector machine. 

Within this rough taxonomy of methods ours directly combines CNN and CRF based learning into a new end-to-end pipeline. In contrast to \cite{scharstein2007learning,li2008learning,zhang2007estimating} our model uses a dense CRF which is used to directly improve the matching cost volume. It can be effectively characterized as a fusion between the CRF and CNN based approaches and thus inherits the benefits of both. The primary contribution of our work lies in our exposition on how to integrate a trainable CRF into an end-to-end deep stereo matching pipeline. We now provide a brief overview of the CRF model to give a firm base to further discussion. 

\section{Conditional Random Fields}
 CRFs represent a family of graphical models for solving multi-label assignment problems over a field of random variables which are conditioned on observed data. In the context of computer vision, CRFs are useful for assigning labels to each image pixel, conditioned on image pixel data while imposing global constraints such as label smoothness. The field is described in terms of the random variables $\{X_{i}\}$ which are associated with a labeling probability function $P\left(X=x|I\right)$ over the set of possible labels $\mathcal{L}=\{l_1,l_2,\ldots,l_k \}$, where I is the input image data. Given an undirected graph $G=\left(V,E\right) $ in which each vertex is associated with a variable $X_i$ and each edge is associated with a joint labeling probability, the joint probability distribution can be written as a Gibbs distribution of the form $P\left(X=x|I\right)=\frac{1}{Z}\exp\left(-E\left(x|I\right)\right)$. We call $E\left(x\right)$ the energy of the configuration $x\in\mathcal{L}$ and $Z$ the partition function, which can be interpreted as a statistical normalization factor over all possible states. For notational convenience we will drop the explicit conditioning on $I$. The MAP estimator for this distribution $\hat{l}=\underset{l}{\operatorname{argmax}} P\left(X=x\right)$ would be the labeling configuration that globally minimizes the energy $E\left(x\right)$. In the CRF model the energy function for a label assignment is given by
\begin{equation}
E\left(x\right)=\sum\limits_i {{\psi _u}\left( {{x_i}} \right) + \sum\limits_{i < j} {{\psi _p}\left( {{x_i},{x_j}} \right)} } 
\end{equation}
where the unary terms $\psi_u\left(x_i\right)$ signify the cost of assigning the label $x_i$ to pixel $i$, and the pairwise terms ${\psi _p}\left( {{x_i},{x_j}} \right)$ signify the joint energy of assigning the labels $\{x_i,x_j\}$ to the pixel pair $(i,j)$. The unary term is set by a per node label probability distribution obtained by a labeling cost function that is dependent on the data. Similarly, the pairwise terms are used to penalize inconsistent labeling for neighboring nodes.

\subsection{Dense CRF with Gaussian potentials} \label{sec:crfprior}
A CRF based on a fully connected graph is called a dense CRF. The dense CRF model imposes global constraints between all the random variables $X_i$ and is therefore expected to yield better results than a model based solely on local connections. The dense CRF optimization however is in general computationally expensive and was deemed impractical for many years until KrÃ€henbÃŒhl \etal  \cite{krahenbuhl2011efficient} showed that using Gaussian edge-weights for inter pixel interactions in a dense CRF enabled a linear time solution. The researchers used a mean field approximation to the CRF distribution, which substitutes the exact distribution $P\left(X\right)$ with a distribution $Q\left(X\right)$ such that the KL-divergence $D\left(P||Q\right)$ is minimal among all possible distributions $Q\left(X\right)$ and that can be expressed as $Q\left( X \right) = \prod\limits_i {Q_i}\left( {X_i} \right) $. This minimization is then performed iteratively using a message passing scheme and can be computed efficiently using high dimensional filtering schemes such as \cite{adams2010fast}. For our CRF model we use the same Gaussian potentials as proposed in \cite{krahenbuhl2011efficient}:
\begin{equation}
{\psi _p}\left( {{x_i},{x_j}} \right) = \mu \left( {{x_i},{x_j}} \right)
\underbrace {\sum\limits_{m = 1}^k {{w^{\left( m \right)}}{k^{\left( m \right)}}
\left( {{f_i},{f_j}} \right)} }_{k\left( {{f_i},{f_j}} \right)},
\end{equation} 
where each $k^{\left(m\right)}$ is a Gaussian kernel, $w^{\left(m\right)}$ is a weight for each kernel, $\mu\left(x_i,x_j\right)$ is a labeling compatibility function and $f_i , f_j$ are feature vectors extracted from the image at locations $i,j$. The label compatibility function $\mu\left(x_i,x_j\right)$ controls the likelihood of two neighboring pixels $x_i,x_j$ to receive a specific combination of labels. Our pairwise terms consist of two Gaussian kernels defined as
\begin{multline}
k\left( {{f_i},{f_j}} \right) = {w^{\left( 1 \right)}}\underbrace {\exp \left( {\frac{{ - {{\left| {{p_i} - {p_j}} \right|}^2}}}{{2\theta _\alpha ^2}}
- \frac{{ - {{\left| {{I_i} - {I_j}} \right|}^2}}}{{2\theta _\beta ^2}}}
\right)}_{appearance\ kernel} \cr
+ {w^{\left( 2 \right)}}\underbrace {\exp \left( {\frac{{ - {{\left| {{p_i} - {p_j}} \right|}^2}}}
{{2\theta _\gamma ^2}}} \right)}_{smoothness\ kernel},
\label{eq:pairwise potentials}
\end{multline}
where $(p_i,p_j)$ are the image coordinates at point $(i,j)$ and $(I_i,I_j)$ are the image RGB coordinates at point $(i,j)$. $\theta_\alpha,\theta_\beta$ and $ \theta_\gamma$ control the width of the Gaussian for each kernel. These kernels are in fact a Bilateral kernel and a spatial Gaussian kernel respectively. The minimization of the KL-divergence $D\left(P||Q\right)$ can be achieved by an iterative update rule written as
\begin{equation}
\begin{multlined}
{Q_i}\left( {{x_i} = l} \right) = \frac{1}{{{z_i}}}\exp \Biggl\{ - {\psi _u}\left( {{ x_i}} \right)-\\
\sum\limits_{l' \in ({\cal L})} {\mu \left( {l,l'} \right)\sum\limits_{m = 1}^k {{w^{\left( m \right)}}\sum\limits_{j \ne i} {{k^{\left( m \right)}}\left( {{f_i},{f_j}} \right){Q_j}\left( {l'} \right)} } } \Biggl\}
\end{multlined}
\label{eq:iterative update}
\end{equation}
as shown in \cite{krahenbuhl2011efficient}.

\subsection{Dense CRF for stereo}
CRF methods are typically used as regularizers which impose constraints onto noisy measurements. This makes them naturally useful for the stereo matching problem in which there are noisy correlations between left and right images. We must then choose the best correlation in a way that satisfies the observed data and the piece-wise smoothness prior. It is both interesting and instructive to note that the popular Semi Global Matching (SGM) \cite{hirschmuller2008stereo} method is in fact an example of a CRF approximation with semi-global constraints which is solved efficiently by dynamic programming. We instead use the fully global dense CRF with Gaussian weights from \cite{krahenbuhl2011efficient} described in Section \ref{sec:crfprior}. To do so we will derive the gradient in Section \ref{end-to-end} and use the method first proposed in \cite{CRFRNN} to propagate error gradients. This is enabled by the iterative update scheme from equation \ref{eq:iterative update} which can be broken down into five steps that require SoftMax, convolution with Gaussian filters and other basic operations which can back propagate their derivatives. The authors of \cite{CRFRNN} show how these operations can be performed in the context of a recurrent neural network (CRF-as-RNN) thereby allowing back-propagation to update the parameters of the CRF. In our work we directly embed the full MC-CNN stereo matching network of \cite{zbontar2015computing} together with the dense CRF layer. During the backward pass the CRF layer propagates it's gradient updates back through the Siamese CNN layers to fine tune those weights. 

\subsection{Stereo matching network}

We make use of the stereo matching cost convolutional network (MC-CNN) proposed by \cite{zbontar2015computing}. In this work we chose the fast version for our experimentation, although the same methodology can be applied to the accurate version as well.

MC-CNN is a patch matching network with a Siamese architecture. The network architecture is described in \cite{zbontar2015computing}. In short it consists of several convolutional and ReLU layers which when applied to a patch output a patch descriptor. The patch descriptor is fed into a Stereo-Join layer which computes the inner products between the descriptors of all the possible matches. The network is trained with a hinge-loss function which given examples of correct and incorrect matches, learns to distinguish between them.

The final output of the network is the matching probability for each patch and it's possible matches. The probability is obtained by calculating the inner product between the patch descriptors at the output of the network. The probabilities are stored in memory as a $3D$ data structure which we call the cost volume. This cost volume contains the matching probability for every possible match and is the input to the CRF stage of our end-to-end network.

In contrast to our network, the MC-CNN is trained solely on patch examples and is therefore a purely local method. As such, it produces a noisy result which requires further regularization. A noisy estimation of the disparity map can be directly obtained by using the $max$ operation on the network output. The refinement step is preformed directly on the cost volume further down the pipeline.


\section{Parameter updates}
The dense CRF approach with Gaussian potentials is a flexible model which is able to impose various smoothness constraints. The constraints are controlled by a number of parameters which should be calibrated in order to obtain good priors. We choose to learn the best parameters using a training set with ground truth disparity from the KITTI autonomous driving dataset \cite{Menze2015CVPR} and from the FlyingThings3D dataset \cite{MIFDB16}.  The parameters which control the CRF are: three Gaussian widths $\theta_{\alpha},\theta_{\beta},\theta_{\gamma}$, the appearance kernel weights $\hat{w}^{\left(1\right)} \in \mathbb{R}^{d_{max}}$, the spatial kernel weights $\hat{w}^{\left(2\right)} \in \mathbb{R}^{d_{max}}$, and the compatibility function weights $\mu \in \mathbb{R}^{d_{max} \times d_{max}}$. According to \cite{krahenbuhl2011efficient} the Gaussian width parameters $\theta_{\alpha},\theta_{\beta},\theta_{\gamma}$ cannot be learned efficiently through gradient methods, and we therefore resolve to learn them using the Nelder-Mead (NM) downhill simplex method \cite{nelder1965simplex} which is a non-gradient based optimization technique. Before we fine tune our full end-to-end MC_CNN-CRF network we first use the NM optimization stage to minimize the average 3-pixel error over a small subset of image pairs from the training data. NM minimizes the function over the space of these variables by evaluating the function at various points which is a computationally reasonable action for this small set of variables but would be infeasible for more. After the minimization is performed we obtain the Gaussian width parameters which we use throughout this work. Additionally we obtain a very rough estimation of the Gaussian kernel weights which act as a good initialization for the gradient based learning process. In fact the kernel weights are integrated into the network through linear operations which are trivially differentiable and make them highly amenable to gradient descent updates as described in \cite{CRFRNN}.  However, in contrast to \cite{CRFRNN} who formulate the multiplication of the kernel weights as a linear layer by extending the weights to a full $d_{max} \times d_{max}$ matrix, we formulate this operation as a convolution layer with a $1\times1$ kernel and $d_{max}$ input and output channels. This allows us to realize the network in a manner that follows \cite{krahenbuhl2011efficient} more accurately without extraneous parameters. 

\subsection{End-to-end learning} \label{end-to-end}
In order to fuse the MC-CNN and CRF stages into one end-to-end network we need to fully define our CRF module in terms of how it acts in a network forward pass as well as its backward pass gradients so that we can use it during training with back-propagation. The output of the MC-CNN network is a cost volume which we consider to be noisy. The function of the CRF module is to reduce the noisiness in the disparity cost matching volume so as to improve the final matching result. In practice this means feeding the full cost volume into the CRF module and producing a cost volume as the output. To do so, we unroll the CRF meanfield loop and convert each step into a neural network action in the same manner as \cite{CRFRNN}. Although we mostly follow their approach, the specific details of our method require non-trivial adaptation. The final cost volume produced as output by the MC-CNN network is computed as
\begin{equation}
{C_{net}}\left( {p,d} \right) = {{\cal P}^L}\left( p \right) \cdot {{\cal P}^R}\left( {p - d} \right),
\label{equation:cnn matching cost}
\end{equation}
where $p$ is a pixel coordinate and $d$ is a disparity value. The depth of the volume is simply the number of disparity values we choose to test over. Therefore ${C_{net}}\left( {p,d} \right)$ is the matching cost for pixel $p$ and disparity $d$ which is produced by the network.  ${\cal P}^L$ and ${\cal P}^R$ are the descriptors produced for the patches at points $p$ and $p-d$ taken from the left and right images respectively and these are the outputs of the Siamese network which forms the core of MC-CNN. In order to attach the MC-CNN and CRF networks the cost volume output ${C_{net}}$  must become the input volume into the CRF. This implies that the operation described in Equation \ref{equation:cnn matching cost} must be defined as a network layer with backward propagation of gradients. Recently, \cite{Luo_2016_CVPR} have described a similar layer which performs forward and backward propagation for the dot product operation between a single patch and it's candidate matches. In our case however, since we are performing global optimization on entire images, we require this layer to produce the backward gradients for all inner products between all the image patches and their matching candidates. This operation is best described as a linear operation 
\begin{equation}
\hat C^{L}\left( p \right) = \left( {\begin{array}{*{20}{c}}
{{{\cal P}^L}\left( p \right)}\\
{{{\cal P}^L}\left( {p + 1} \right)}\\
\vdots \\
{{{\cal P}^L}\left( {p + d} \right)}
\end{array}} \right){{\cal P}^R}\left( p \right).
\end{equation}
By treating the left image representation $\mathcal{P}^{L}$ as constants, it is easy to see that the adjoint operator required for gradient back propagation is the transpose of the matrix. By this reasoning we obtain the following rule for gradient propagation
\begin{equation}
\hat G_{in}^L\left( p \right) = {\left( {\begin{array}{*{20}{c}}
{{{\cal P}^L}\left( p \right)}\\
{{{\cal P}^L}\left( {p + 1} \right)}\\
\vdots \\
{{{\cal P}^L}\left( {p + d} \right)}
\end{array}} \right)^T}\hat G_{out}^R\left( p \right).
\end{equation}
Here $G_{in}$ and $G_{out}$ represent the input and output gradients respectively. The adjoint operator used for the gradient propagation to the right image descriptors is attained following the same logic such that
\begin{equation}
\hat G_{in}^R\left( p \right) = {\left( {\begin{array}{*{20}{c}}
{{{\cal P}^R}\left( p \right)}\\
{{{\cal P}^R}\left( {p - 1} \right)}\\
\vdots \\
{{{\cal P}^R}\left( {p - d} \right)}
\end{array}} \right)^T}\left( {\begin{array}{*{20}{c}}
{G_{out}^L\left( {p,0} \right)}\\
{G_{out}^L\left( {p - 1,1} \right)}\\
\vdots \\
{G_{out}^L\left( {p - d,d} \right)}
\end{array}} \right).
\end{equation}
This scheme allows us to add the necessary gradient computation mechanism to fuse the MC-CNN network to the CRF network and train them jointly using whole images as input and disparity probabilities as the output.

\subsection{Training loss}
We experimented with several loss functions such as MSE, KL-Divergence, Cross Entropy Loss and Piecewise linear loss \cite{levi2015stixelnet}. From these loss functions the Multi-class Cross Entropy and Piecewise linear losses, which are applied to each pixel location separately, provided the best convergence. The PL loss is an extension of Cross Entropy which performs a linear interpolation between the discrete bins, allowing real numbers as labels. During our training attempts the network exhibited signs of lowering the loss rate while increasing the $L1$ error relative to the ground truth. Although at first this appeared to be a symptom of overfitting, it became apparent that the loss function was encouraging the probability of correct labels to rise, but was not penalizing the bins of the incorrect labels. To remedy this we introduced an entropy penalty into our cost function which pushes the network output towards lower entropy. We also tested the $L1$ penalty but found it to be less effective. 
The entropy penalty that we introduce is of the form $P =  - \sum\limits_{i = 1}^{{d_{\max }}} {\alpha {p_i}\log \left( {{p_i}} \right)}$ and it's gradient is of the form $p_i^' =  - \alpha \left( {1 + \log \left( {{p_i}} \right)} \right)$. By adding this gradient to the loss gradients we achieve a more stable gradient descent which converges both in the loss rate and in the $L1$ error domains. During training we use a penalty factor of $\alpha=0.1$. 

\subsection{Implementation details} \label{section:implementation}
We use the open source code for the fast architecture for MC-CNN \cite{zbontar2015computing} made available by the authors. We then implemented the dense CRF network layers within the Torch and CUDA frameworks. We trained our network on an Nvidia TITAN-X GPU. The code for efficient Gaussian filtering on GPU was adopted from the implementations by \cite{adams2010fast} and \cite{CRFRNN}. The CRF network is attached to the output of the cost network via the inner-product layer described in section \ref{end-to-end}. We use training sets of $160$ stereo pairs out of the KITTI 2015 \cite{Menze2015CVPR} and approximately $15k$ examples from FlyingThings3D \cite{MIFDB16} datasets to train our network. We initialize our CRF to the parameters calibrated by the Nelder-Mead simplex method \cite{nelder1965simplex}. This is performed as a training step by evaluating the 3-pixel error over a subset of $20$ images out of the training set, over the space of five tunable parameters. The parameters which are learned are $\theta_{\alpha} = 18.65$, $\theta_{\beta} = 4.39$ and $\theta_{\gamma} = 2.13$. The Gaussian filter weights are $\hat{w}^{\left(1\right)}=18.68$ and $\hat{w}^{\left(2\right)}=68.68$. We also initialize the compatibility function according to the following rule
\begin{equation}
\mu \left( {i,j} \right) = \left\{ {\begin{array}{*{20}{c}}
{ - 1 + 0.2\left| {i - j} \right|}&{\left| {i - j} \right| \le 4}\\
0&{otherwise}
\end{array}} \right.
\label{eq:compatibility}
\end{equation}

\begin{figure}[h]
\begin{center}

\includegraphics[width=0.48\linewidth]{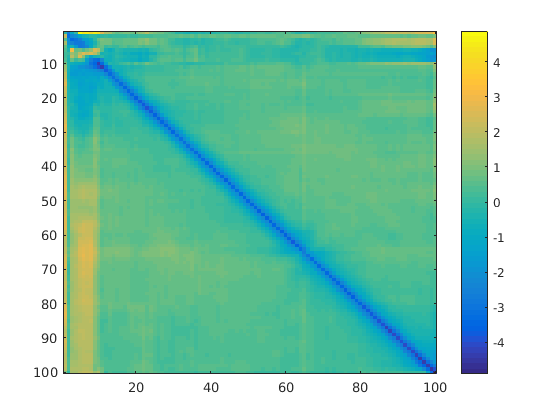}
\includegraphics[width=0.48\linewidth]{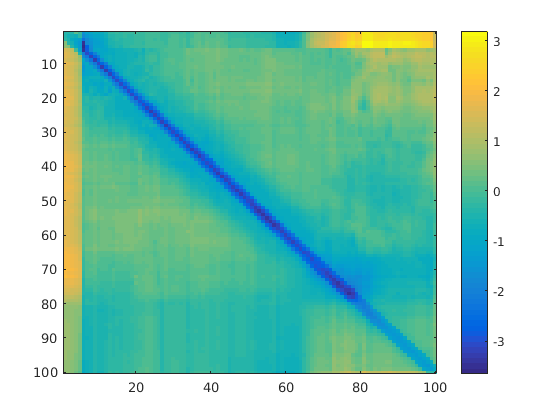}

\caption{ The final trained compatibility weights $\mu \left( {i,j} \right)$ are depicted. This diagonal structure which arises from the training depicts the learned CRF smoothness priors. On the left: weights trained from FlyingThings3D\cite{MIFDB16}. On the right: weights trained from KITTI \cite{Menze2015CVPR}}
\label{fig:compat_weights}
\end{center}
\end{figure}

The values of $\mu \left( {i,j} \right) $ are displayed in Figure \ref{fig:compat_weights}. The diagonal structure which arises from the learning process depicts the spatial smoothness prior which is imposed by the CRF, where nearby labels have high probability of receiving similar disparities. It is clear that the weights obtained on the KITTI dataset are not fully formed, specifically in the larger disparity region. We discuss this further in Section \ref{sec:conclusions}

The rule in equation \ref{eq:compatibility} encodes the intuition that neighboring pixels are most likely to receive the same labels, and as the label difference becomes larger the likelihood declines linearly until it plateaus at $0$ at a distance of $4$. This provides a good approximation for a piece-wise smooth labeling prior.  At each training iteration the network receives a stereo image pair from the training set, along with it's corresponding ground truth disparity map. The disparity maps in KITTI are sparse and we therefore must take this into consideration. We create a mask for each ground truth image of the empty disparity locations, and use it to compute the loss function for the allowed positions only. Additionally, we exclude from the mask any occluded and over exposed pixels which might hinder the learning process. During the backward pass of the network we zero the gradients which emanate from the data-less regions and propagate only the useful gradients down the network.  We found that the best results are gained using Adagrad optimization \cite{duchi2011adaptive}. Initially we train only the CRF weights for 30 epochs while freezing the rest of the network, and later we release the weights in both parts of the network and allow them to learn jointly. We use an initial learning rate of $0.1$ for the CRF weights and relay on Adagrad to set the learning rate appropriately. We train the joint network for an additional 50 epochs. 

\section{Experiments}
\begin{table*}[!h]
\centering

\newcolumntype{P}[1]{>{\centering\arraybackslash}p{#1}}
\newcolumntype{M}[1]{>{\centering\arraybackslash}m{#1}}

\begin{tabular}{@{}l|P{1cm}P{1cm}|P{1cm}P{1cm}|P{1cm}P{1cm}|P{1cm}P{1cm}l@{}}
\toprule
Pipeline & \multicolumn{4}{c|}{KITTI 2015 \cite{Menze2015CVPR}} & \multicolumn{4}{c}{FlyingThings3D \cite{MIFDB16}}& \\ \midrule

& \multicolumn{2}{c|}{post process} & \multicolumn{2}{c|}{no post p.} & \multicolumn{2}{c|}{post process} & \multicolumn{2}{c}{no post p.}   \\
& 3-pixel & 1-pixel & 3-pixel & 1-pixel & 3-pixel & 1-pixel & 3-pixel & 1-pixel  \\ \midrule

{\cite{zbontar2015computing}} {MC-CNN --- SGM} 
&\bf{3.78} & 12.73 &  \bf{4.74}  &14.45 & 9.66 & 11.96  & 14.01 &  15.88&    \\

{(ours)} {MC-CNN --- CRF}
& 5.12 & 12.55 &  6.54  & 15  & \bf{8.85}  & \bf{10.98} &  12.1   &  13.88 &   \\

{(ours)} {MC-CNN --- CRF --- SGM}
&4.7  & \bf{11.74} &  5.32 &  \bf{13.03} & \bf{8.85} &  11.02 & 10.88  & \bf{12.64}  & \\

{\cite{MIFDB16}} {Mayer \etal}
&-- &  -- &  9.97 & 31.36  & --  &-- & \bf{9.12}  & 14.21 & \\
\end{tabular}
\caption {3-pixel and 1-pixel error rates shown for our method compared to \cite{zbontar2015computing,MIFDB16}. The post processing methods we used were taken from \cite{zbontar2015computing}. }
\label{tabel:results}
\end{table*}

We conducted our evaluation on the KITTI 2015 \cite{Menze2015ISA,Menze2015CVPR} benchmark as well as on the FlyingThings3D \cite{MIFDB16} synthetic dataset. For both we set aside validation sets and trained on the remaining pairs. For KITTI we used $40$ pairs for validation and in FlyingThings3D we used $100$ pairs. We trained MC-CNN \cite{zbontar2015computing} on the training set alone and then trained our CRF model as described in Section \ref{section:implementation}. 
After the training process we tested the average 3-pixel error rate on the validation set using several configurations:
\begin{itemize}
\item MC-CNN followed by SGM as described by \cite{zbontar2015computing}
\item MC-CNN followed by CRF in place of SGM
\item MC-CNN followed by CRF and then SGM
\item DispNetCorr1D as described by \cite{MIFDB16}
\end{itemize}
Each configuration was tested with and without the post processing methods described by \cite{zbontar2015computing}. 

\section{Results}

\begin{figure*}[p]
\centering
\begin{subfigure}{.5\textwidth}
  \centering
  \includegraphics[width=1\linewidth, height=0.14\textheight]{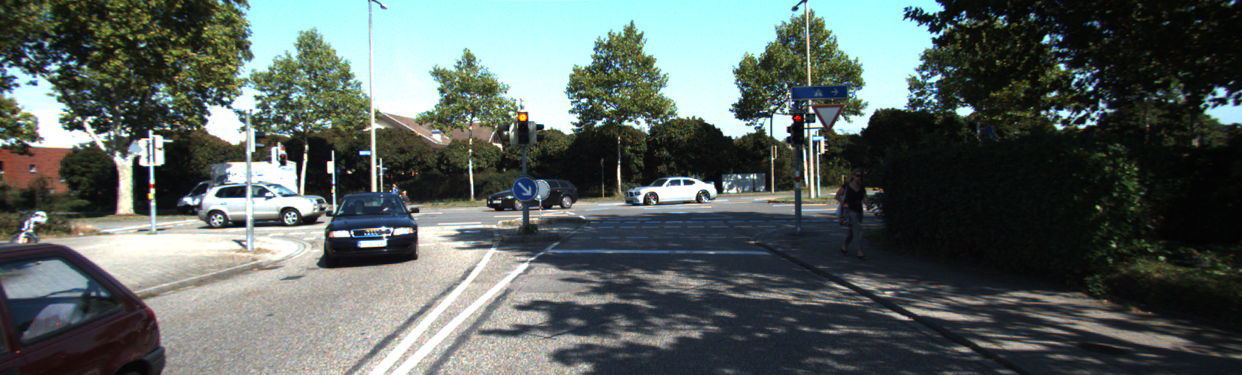}
  \caption{Input image}
  \label{fig:sub1}
\end{subfigure}%
\begin{subfigure}{.5\textwidth}
  \centering
    \includegraphics[width=1\linewidth, height=0.14\textheight,,trim={0 2cm 0 0},clip]{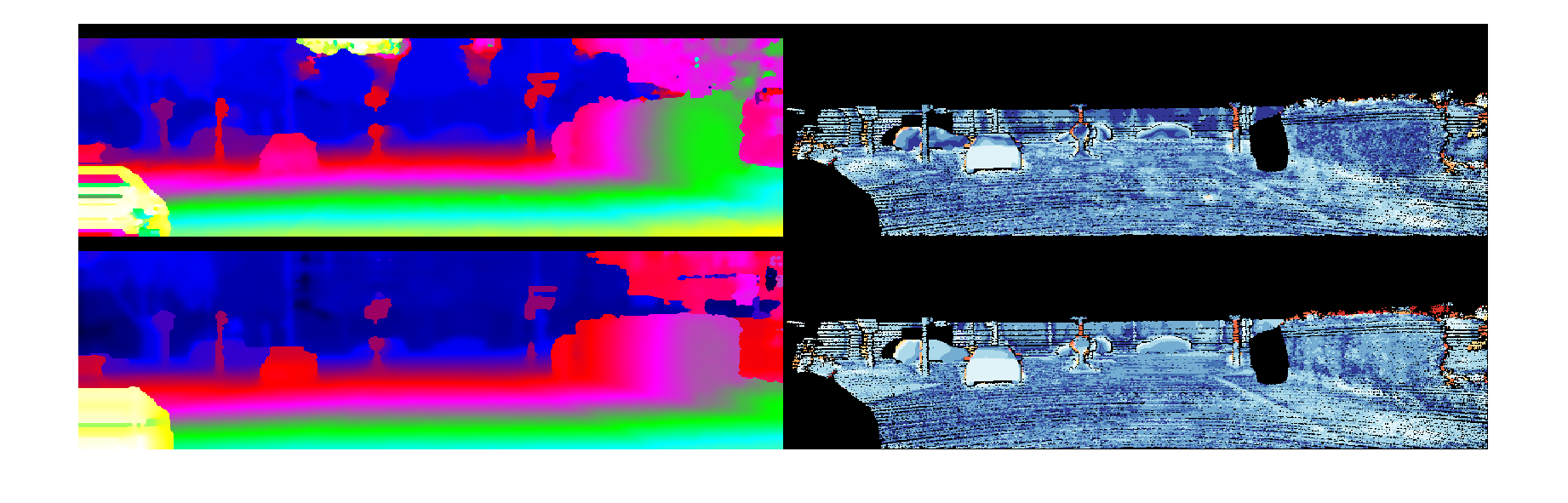}
  \caption{Top: our method - $1.09\%$ error rate Bottom: \cite{zbontar2015computing} method - $1.45\%$ error rate}
  \label{fig:sub2}
\end{subfigure}

\begin{subfigure}{.5\textwidth}
  \centering
  \includegraphics[width=1\linewidth, height=0.14\textheight]{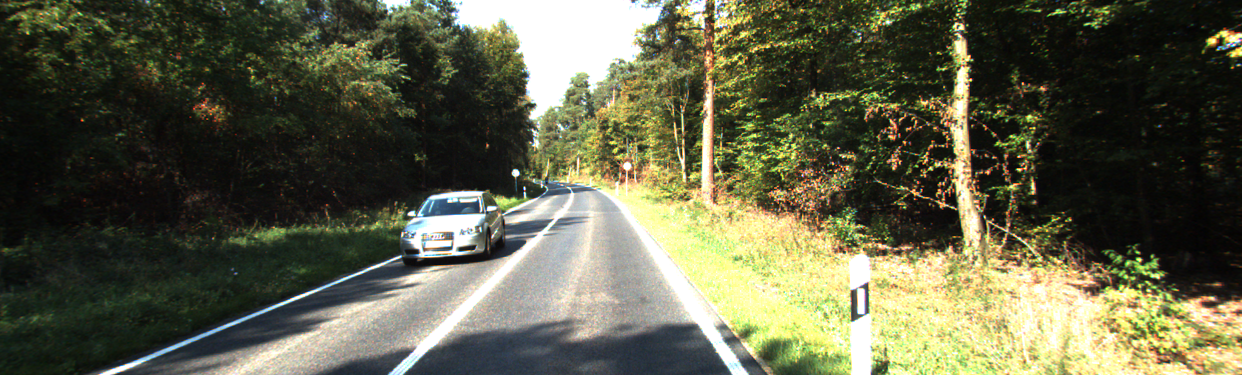}
  \caption{Input image}
  \label{fig:sub1}
\end{subfigure}%
\begin{subfigure}{.5\textwidth}
  \centering
    \includegraphics[width=1\linewidth, height=0.14\textheight,,trim={0 2cm 0 0},clip]{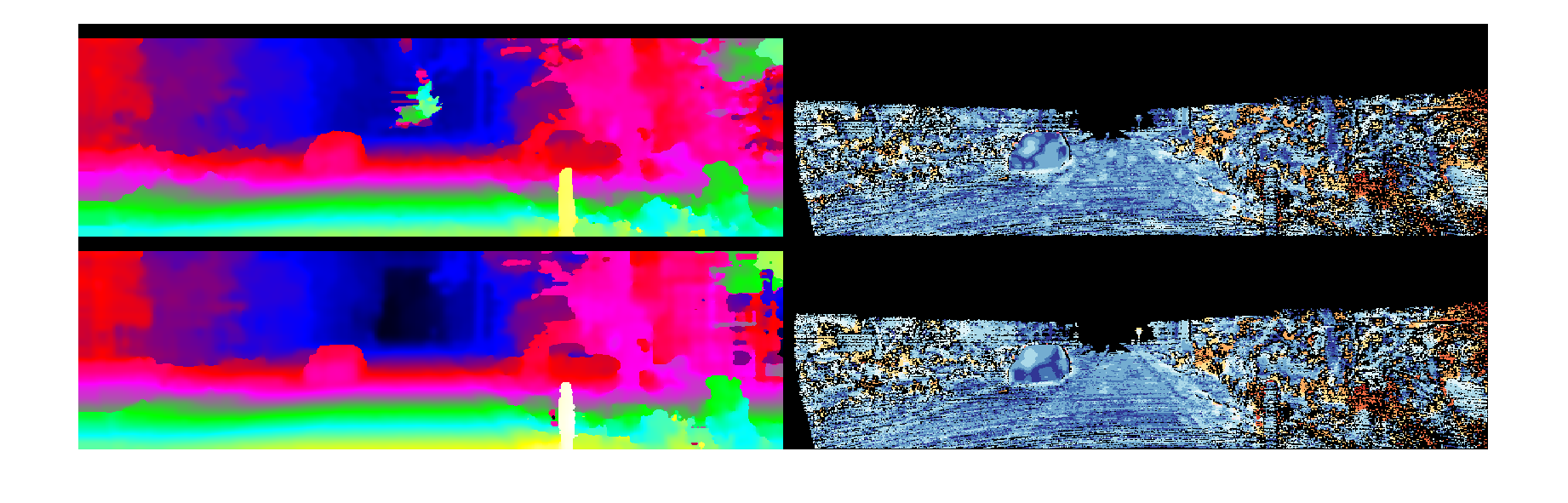}
  \caption{Top: our method - $6.37\%$ error rate  Bottom: \cite{zbontar2015computing} method - $6.99\%$ error rate}
  \label{fig:sub2}
\end{subfigure}

\caption{This figure displays an example of a scene from KITTI 2015 where our method achieves a {\bf lower error rate} than that of \cite{zbontar2015computing}. On the left side are the input images and on the right are the computed disparity maps and error maps. Here it we demonstrate the advantages of our method. The disparity map is produced with finer detail and without the streaking artifacts which happen when using SGM.}
\label{fig:examples_pos}
\end{figure*}

\begin{figure*}
\centering
\begin{subfigure}{.5\textwidth}
  \centering
  \includegraphics[width=1\linewidth, height=0.14\textheight]{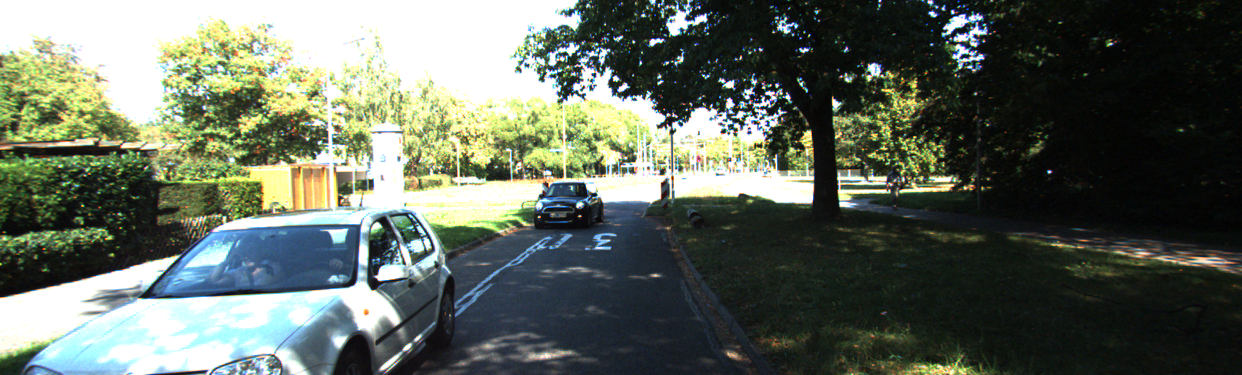}
  \caption{Input image}
  \label{fig:sub1}
\end{subfigure}%
\begin{subfigure}{.5\textwidth}
  \centering
    \includegraphics[width=1\linewidth, height=0.14\textheight,,trim={0 2cm 0 0},clip]{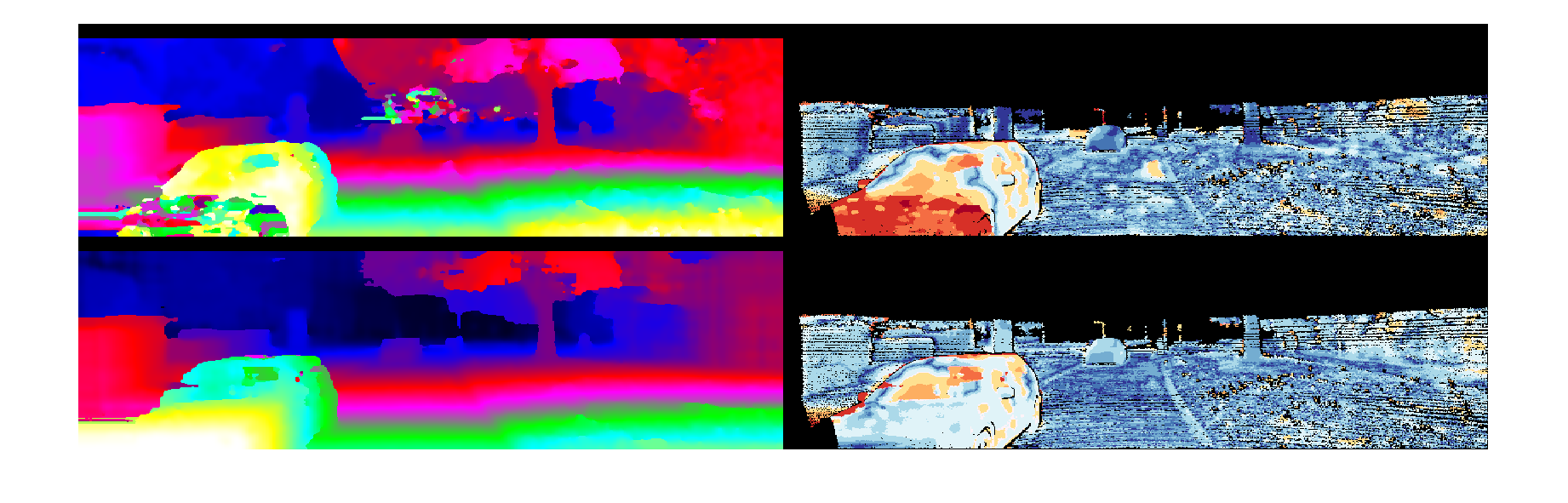}
  \caption{Top: our method - $25.46\%$ error rate  Bottom: \cite{zbontar2015computing} method - $10.36\%$ error rate}
  \label{fig:sub2}
\end{subfigure}

\begin{subfigure}{.5\textwidth}
  \centering
  \includegraphics[width=1\linewidth, height=0.14\textheight]{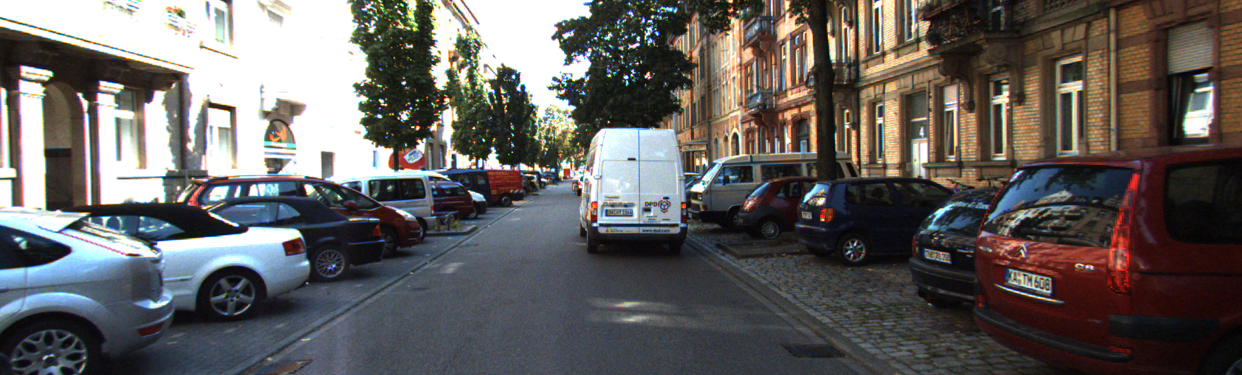}
  \caption{Input image}
  \label{fig:sub1}
\end{subfigure}%
\begin{subfigure}{.5\textwidth}
  \centering
    \includegraphics[width=1\linewidth, height=0.14\textheight,,trim={0 2cm 0 0},clip]{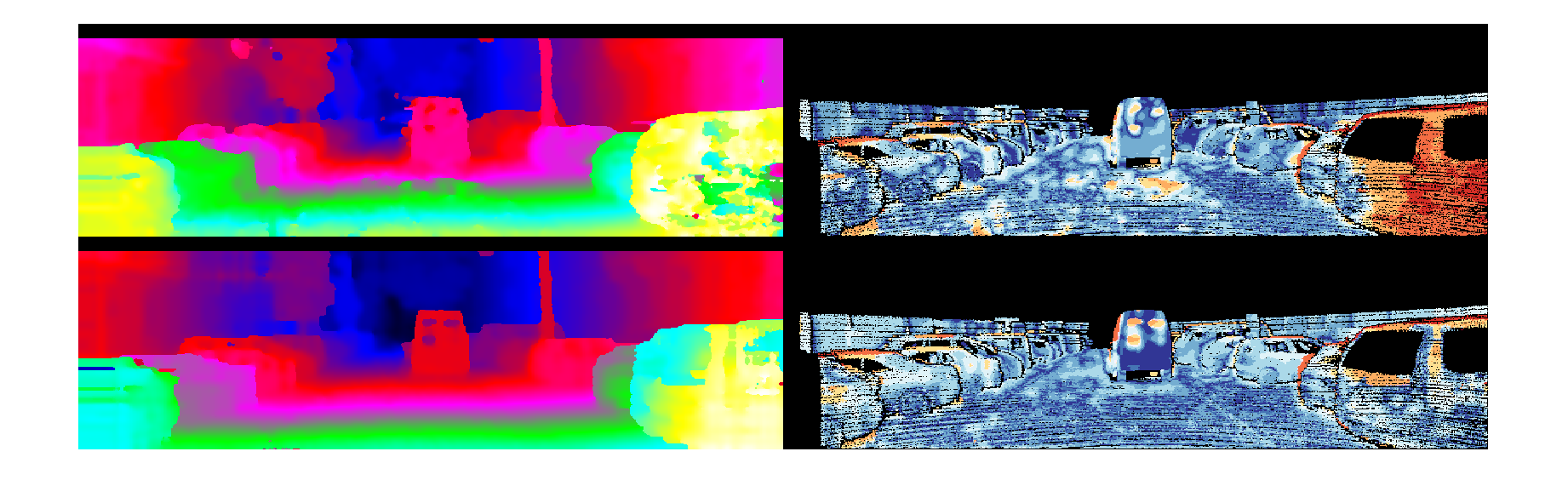}
  \caption{Top: our method - $14.59\%$ error rate  Bottom: \cite{zbontar2015computing} method - $6.59\%$ error rate}
  \label{fig:sub2}
\end{subfigure}

\caption{This figure displays an example of a scene from KITTI 2015 where our method achieves a {\bf higher error rate} than that of \cite{zbontar2015computing}. On the left side are the input images and on the right are the computed disparity maps and error maps. It is clearly visible on these examples that we fail mainly on the near objects (large disparity). This is due to lacking examples for near objects in the KITTI dataset.}
\label{fig:examples_neg}
\end{figure*}

\begin{figure*}
\centering
\begin{subfigure}{.5\textwidth}
  \centering
  \includegraphics[width=1\linewidth, height=0.12\textheight,trim={0 -2.5cm 0 0},clip]{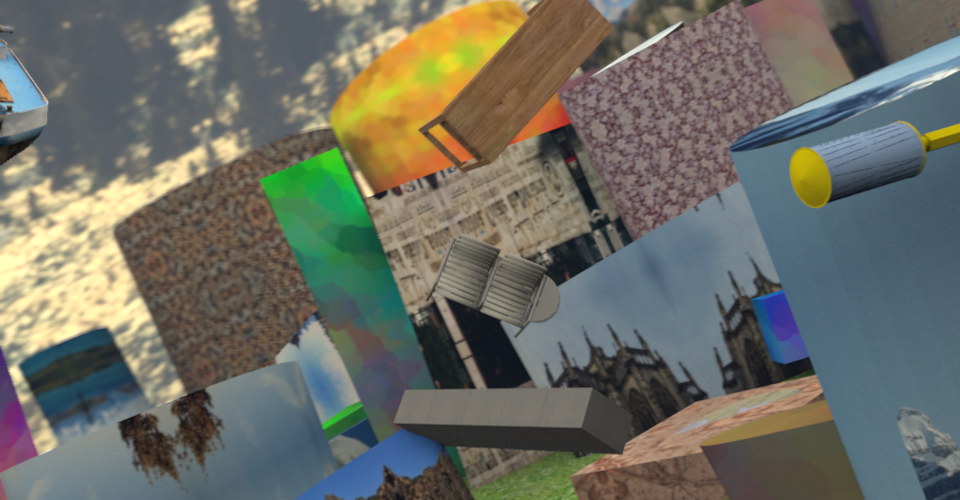}
  \caption{Input image}
  \label{fig:sub1}
\end{subfigure}%
\begin{subfigure}{.5\textwidth}
  \centering
    \includegraphics[width=1\linewidth, height=0.12\textheight,trim={0 0 0 0},clip]{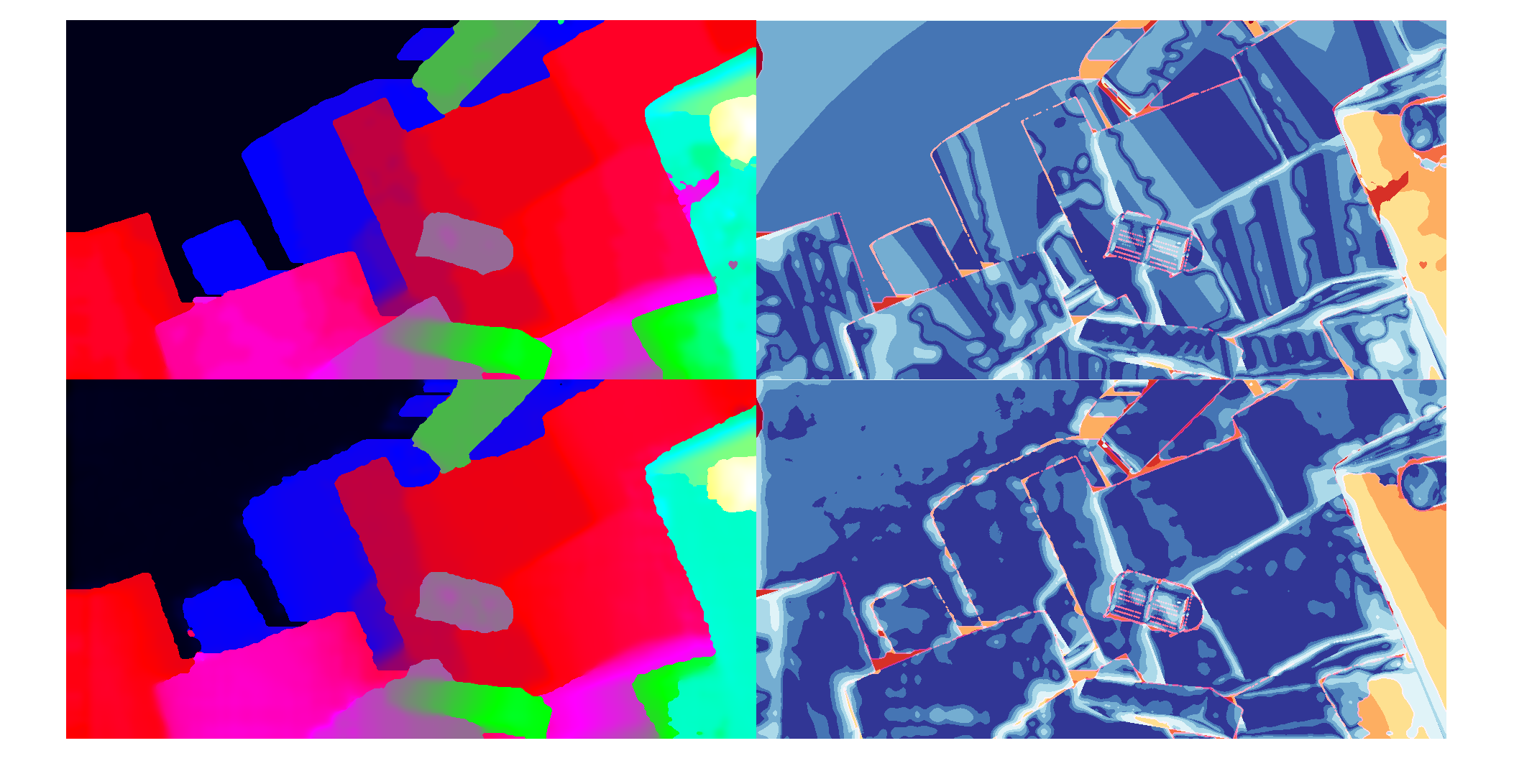}
  \caption{Top: our method - $9.68\%$ error rate  Bottom: \cite{zbontar2015computing} method - $11.85\%$ error rate}
  \label{fig:sub2}
\end{subfigure}

\begin{subfigure}{.5\textwidth}
  \centering
  \includegraphics[width=1\linewidth, height=0.12\textheight,trim={0 -2.5cm 0 0},clip]{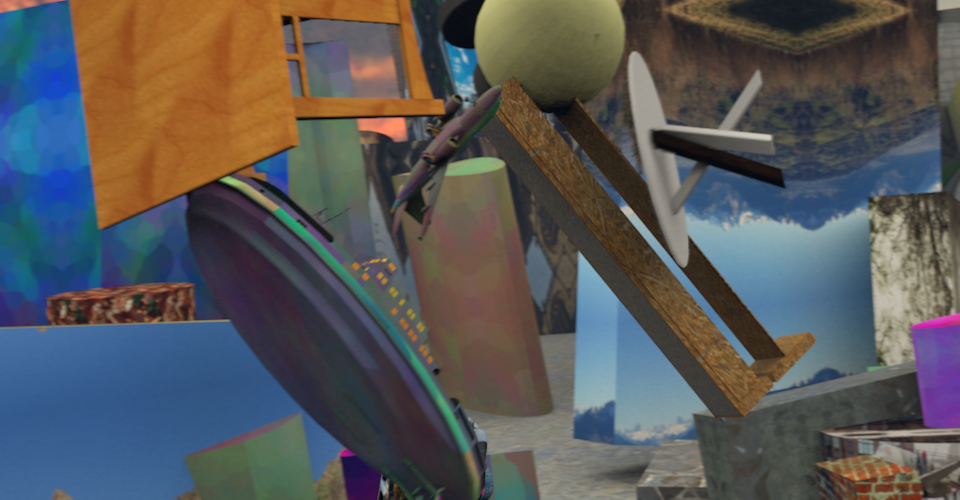}
  \caption{Input image}
  \label{fig:sub1}
\end{subfigure}%
\begin{subfigure}{.5\textwidth}
  \centering
 \includegraphics[width=1\linewidth, height=0.12\textheight,trim={0 0 0 0},clip]{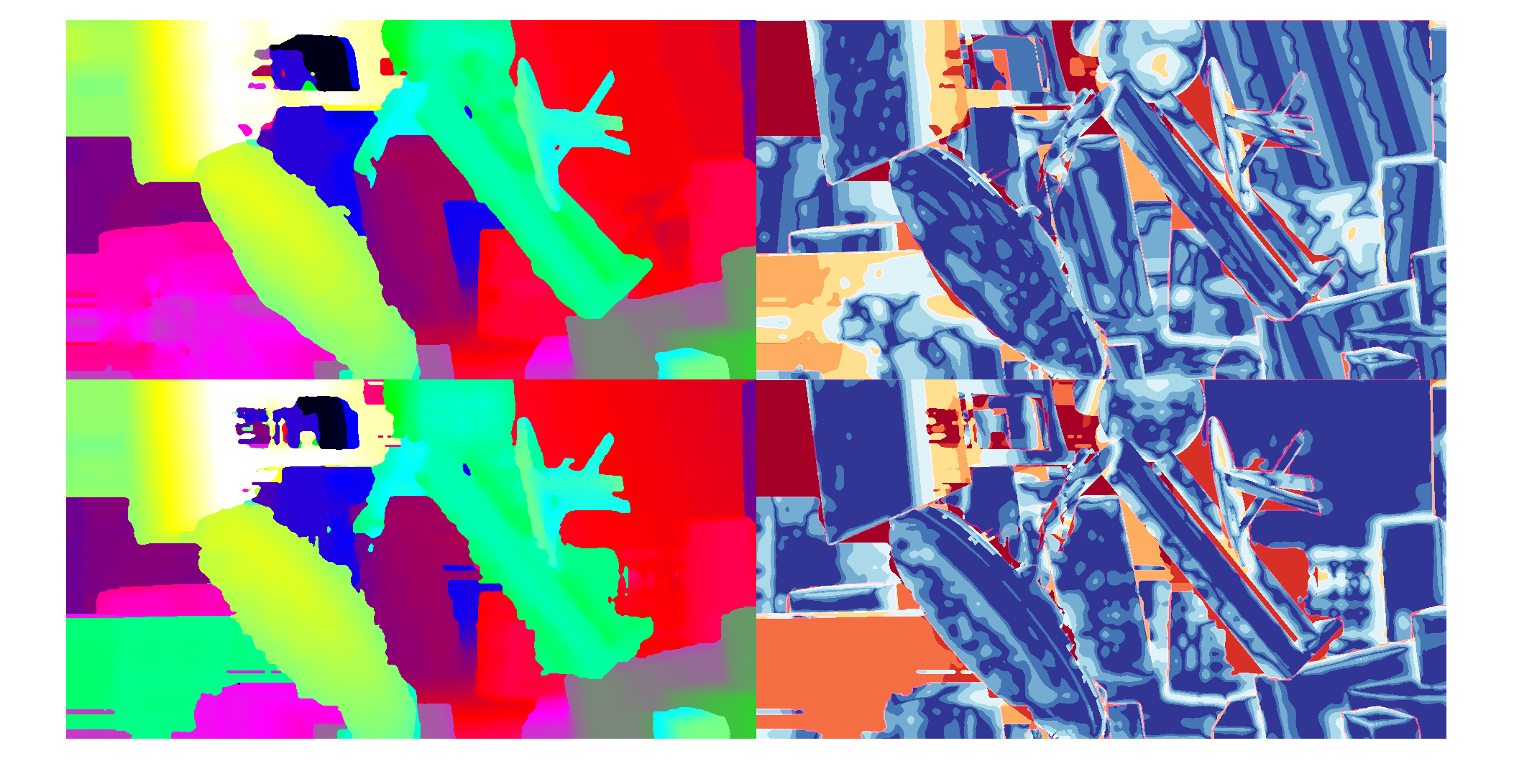}
  \caption{Top: our method - $14.36\%$ error rate  Bottom: \cite{zbontar2015computing} method - $19.95\%$ error rate}
  \label{fig:sub2}
\end{subfigure}

\begin{subfigure}{.5\textwidth}
  \centering
  \includegraphics[width=1\linewidth, height=0.12\textheight,trim={0 -2.5cm 0 0},clip]{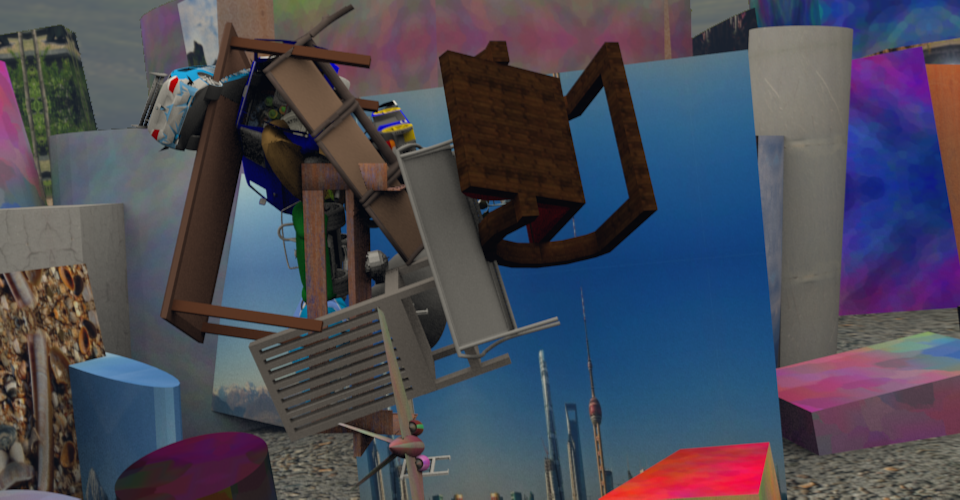}
  \caption{Input image}
  \label{fig:sub1}
\end{subfigure}%
\begin{subfigure}{.5\textwidth}
  \centering
 \includegraphics[width=1\linewidth, height=0.12\textheight,trim={0 0 0 0},clip]{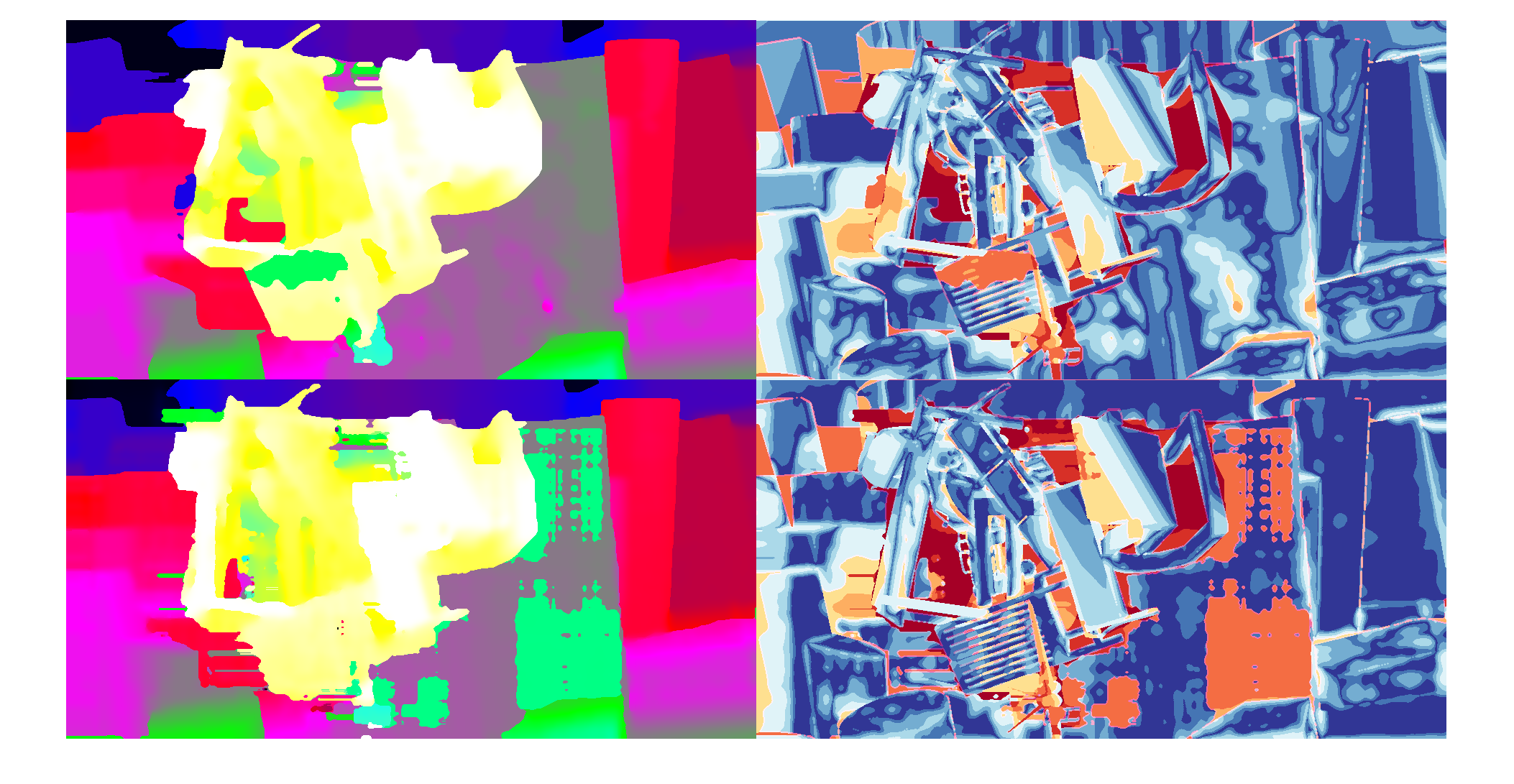}
  \caption{Top: our method - $3.4\%$ error rate  Bottom: \cite{zbontar2015computing} method - $8.36\%$ error rate}
  \label{fig:sub2}
\end{subfigure}

\begin{subfigure}{.5\textwidth}
  \centering
  \includegraphics[width=1\linewidth, height=0.12\textheight,trim={0 -2.5cm 0 0},clip]{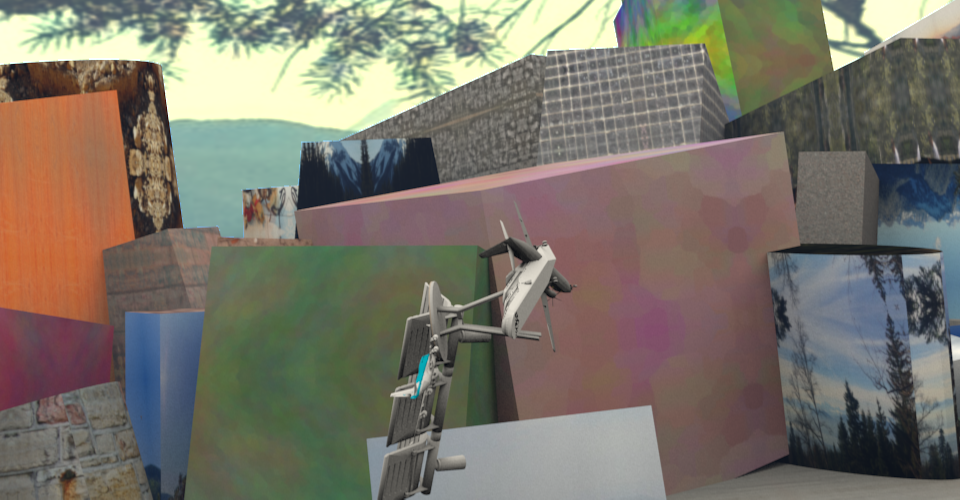}
  \caption{Input image}
  \label{fig:sub1}
\end{subfigure}%
\begin{subfigure}{.5\textwidth}
  \centering
 \includegraphics[width=1\linewidth, height=0.12\textheight,trim={0 0 0 0},clip]{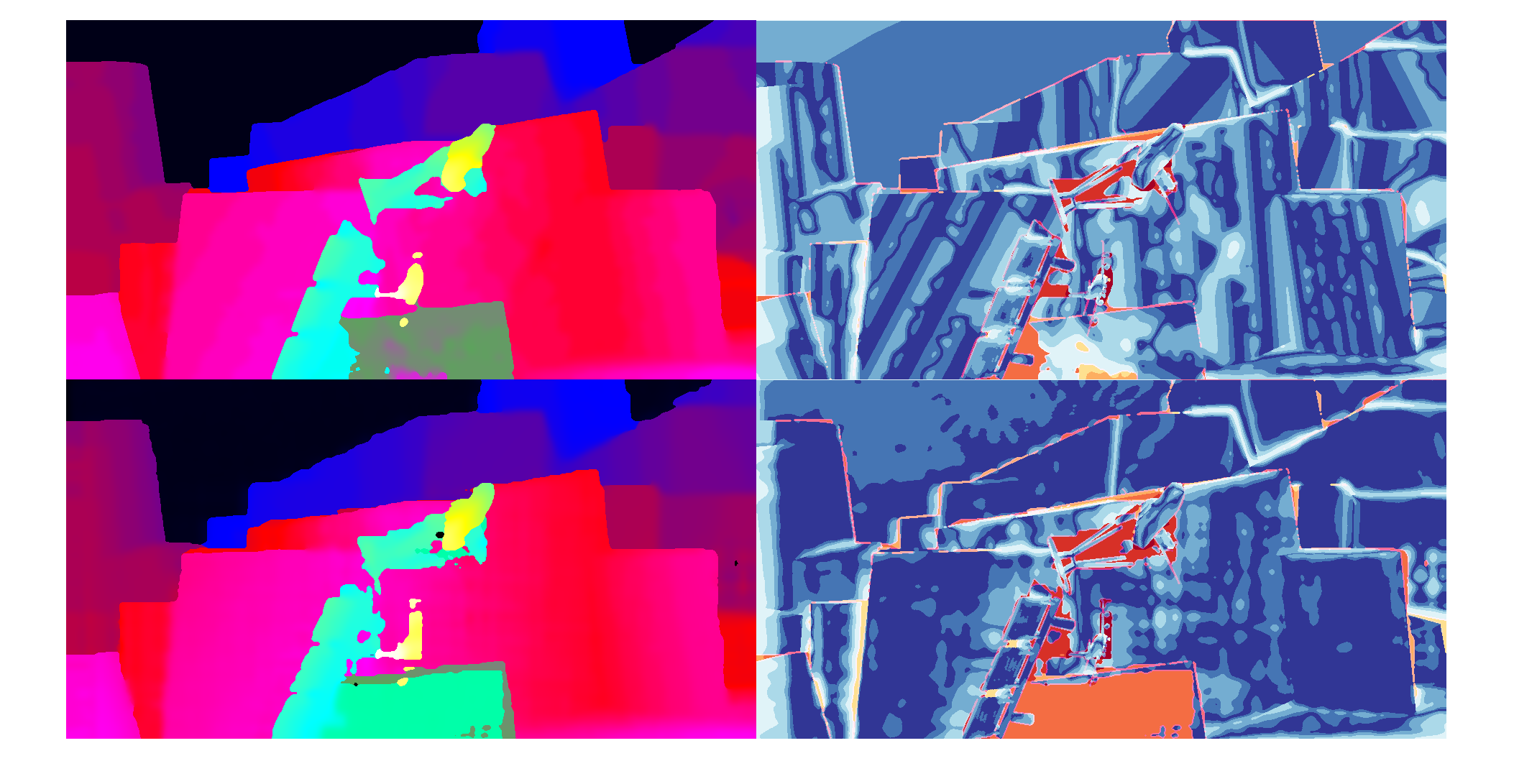}
  \caption{Top: our method - $7.19\%$ error rate  Bottom: \cite{zbontar2015computing} method - $8.48\%$ error rate}
  \label{fig:sub2}
\end{subfigure}

\caption{Examples of disparity prediction with error rates for our method versus \cite{zbontar2015computing} on the FlyingThings3D dataset \cite{MIFDB16}. Here we show that our method indeed beats \cite{zbontar2015computing} in many cases and does not suffer from errors in near disparities as on the KITTI dataset. This is due to the large size of the dataset and the more even distribution of disparities throughout the samples.}
\label{fig:examples_flying}
\end{figure*}

As shown in Table \ref{tabel:results} we are able to outperform both methods when training and testing on the FlyingThings3D dataset. On the KITTI dataset our method outperforms \cite{zbontar2015computing} for $1\%$ error but not for $3\%$. This indicates that our method is generally accurate, however it is affected by areas of high error in some cases. It is our belief that this is due to the fairly small size of the KITTI dataset which contains only $200$ training samples compared to $20k$ samples in FlyingThings3D.  Specifically there appears to be an underepresentation in the number of nearby surfaces in the KITTI scenes, whereas this is not the case for the FlyingThings3D dataset. Since we are training in an end-to-end fashion using whole images as our input data, we are more susceptible to the lack of training data than \cite{zbontar2015computing} who train on image patches. During our work we noticed that using modern optimizer such as Adagrad \cite{duchi2011adaptive} partially remedy this by increasing the learning rate for weights that rarely receive gradients. This training scheme dramatically improves on this issue, but still does not completely solve it in this case. Our results show that most of our errors are concentrated in large disparity regions. Examples of this can be seen in Figure \ref{fig:examples_neg}.  In other examples however we can see that our method is able to perceive finer details and also does not suffer from the streaking artifact of SGM. Examples of this are shown in Figure \ref{fig:examples_pos}. The large disparity problem does not negatively impact our results on the FlyingThings3D dataset. Several representative examples are shown in Figure \ref{fig:examples_flying}. The fact that our proposed method outperforms that of \cite{zbontar2015computing} on both datasets illustrates that including a powerful regularizer into an end-to-end trained system has a positive impact on the quantitative results. This is advantage is further highlighted when a very large dataset is available for training. This indicates that our approach is most suited to be truly beneficial to large datasets with noisy data that have a good coverage of their data space. I addition, our method coupled with SGM can increase the performance of the disparity recovery further by harnessing the advantages of both methods as shown in Table \ref{tabel:results}.

\section{Conclusions}\label{sec:conclusions}

In this work we have described a novel fused CNN and CRF pipeline for stereo reconstruction that permits end-to-end training. Our results indicate that this direction is fruitful and provides a more holistic, less fragmented solution to stereo reconstruction which is a problem that has typically been highly engineered. We show that for the specific task of stereo matching we are able to beat competing CNN based methods if the dataset is large enough. When data is more scarce our approach is still able to achieve very competitive results, especially in $1\%$ error rate. It should be noted that poorly covered characteristics in the data will yield errors in the predicted disparities (Figure \ref{fig:examples_neg}). It is our belief that improving the training methods and data processing, our method can overcome the large disparity issue on KITTI which would greatly improve the results on future datasets. On a more general level we believe that our findings justify advocating to other researchers the benefits of focusing efforts on building specialized network architectures which implement more task specific classical algorithms as a layer in a neural network for end-to-end training. 



{\small
\clearpage
\bibliographystyle{ieee}
\bibliography{egbib}
}

\end{document}